\documentclass{bmvc2k}

\usepackage{amsmath}
\usepackage{amssymb}
\usepackage{eurosym}
\usepackage{calc}
\usepackage{multirow}
\usepackage{adjustbox}
\newcommand{\R}{\mathbb{R}}

\title{Prototypical Priors: From Improving Classification to Zero-Shot Learning}

\addauthor{Saumya Jetley}{sjetley@robots.ox.ac.uk}{1}
\addauthor{Bernardino Romera-Paredes}{bernard@robots.ox.ac.uk}{1}
\addauthor{Sadeep Jayasumana}{sadeep@robots.ox.ac.uk}{1}
\addauthor{Philip Torr}{phst@robots.ox.ac.uk}{1}

\addinstitution{
 University of Oxford\\
 Oxford, UK
}

\runninghead{Jetley et al.}{Prototypical Priors: Classification to Zero-Shot Learning}

\begin{document}

\maketitle

\begin{abstract}
Recent works on zero-shot learning make use of side information such
as visual attributes or natural language semantics to define the relations
between output visual classes and then use these relationships to draw inference
on new unseen classes at test time. In a novel extension to this idea,
we propose the use of visual prototypical concepts as side information. For
most real-world visual object categories, it may be difficult to establish
a unique prototype. However, in cases such as traffic signs, brand
logos, flags, and even natural language characters, these prototypical
templates are available and can be leveraged for an improved recognition
performance. 


The present work proposes a way to incorporate this prototypical information
in a deep learning framework. Using prototypes as prior information, the deepnet pipeline
learns the input image projections into the prototypical embedding space subject
to minimization of the final classification loss. Based on our
experiments with two different datasets of traffic signs and brand
logos, prototypical embeddings incorporated in a conventional convolutional neural network improve
the recognition performance. Recognition accuracy on the Belga logo dataset is
especially noteworthy and establishes a new state-of-the-art. In zero-shot
learning scenarios, the same system can be directly deployed to draw inference on unseen classes by simply adding the prototypical information 
for these new classes at test time. Thus, unlike earlier approaches, testing
on seen and unseen classes is handled using the same pipeline, and the system
can be tuned for a trade-off of seen and unseen class performance as per 
task requirement. Comparison with one of the latest works in the zero-shot
learning domain yields top results on the two datasets mentioned above.

\end{abstract}

\section{Introduction}

Automatic object recognition has witnessed a huge improvement in recent years due to the successful 
application of convolutional neural networks (CNN). This boost in performance can be explained by 
the replacement of heuristic parts in the previous feature representation approaches by a methodology 
\cite{Lee/icml2009,Krizhevsky/nips2012} based on learning the features straight from the data. 
The learned feature representation, which is tailored to the given learning scenario, generally 
outperforms heuristic approaches provided the training data is sufficient.
When learned over a significant sample variety, this representation captures regularities across samples of a class 
that help distinguish it from all the other classes.      
\begin{figure}
\begin{center}
\includegraphics[scale=0.55]{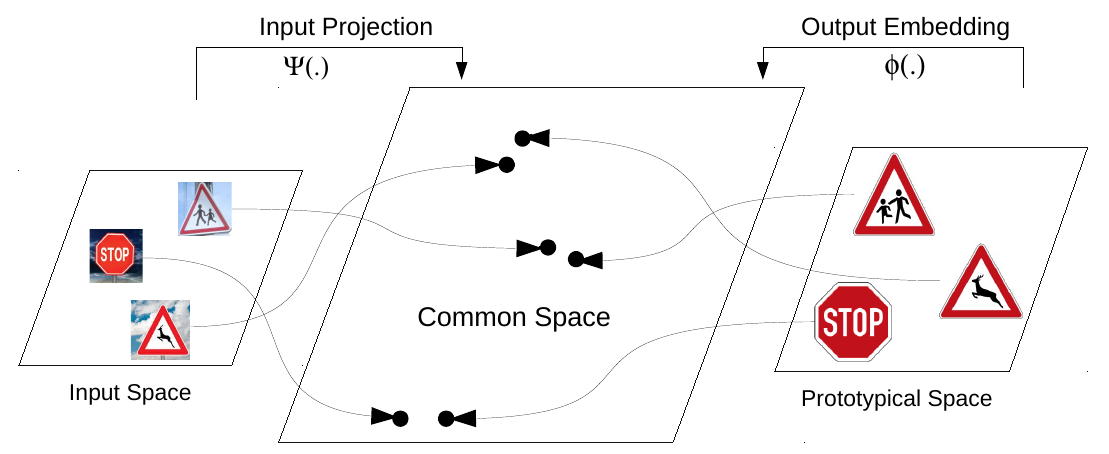}
\end{center}
\protect\caption{\label{fig:scheme} A joint embedding space defined by
the prototypes}
\end{figure}

In an alternative setup, the object recognition problem can be posed as one
in which objects in real images are identified by treating them as imperfect and corrupt copies 
of prototypical concepts. This assumption provides an additional premise that
the different samples of a class are not only similar to each other but also resemble a 
unique prototype. These prototypical concepts are in many cases not
available, for example, there does not exist a chair that contains only the
essence of \textit{chair} and nothing else.
However, there are many scenarios where such prototypical instances do exist. An example of this is traffic 
sign recognition, in which each traffic sign class has its canonical template. Real world images contain 
imperfect instances of it, these imperfections being caused by different viewpoints, light conditions, 
damage to surface, among others. These canonical templates, hereafter
uniformly referred to as prototypes (\textit{an original or first model of
something from which other forms are copied or developed\footnote{Definition taken from Merriam-Webster.com dictionary.%
}}), can play a very important role in recognition.
Conceivably, this prototypical information can benefit by - (a) guiding the learning process and 
(b) establishing an output embedding space where the relationship between
output visual classes can be used to transfer the learned knowledge to unseen
classes directly at test time.

In the present work, we focus on adding this prototypical prior information into convolutional neural networks. 
The underlying idea is that the high-level representation learned by a CNN should be comparable to the 
information extracted from the prototypes. An interpretation of this is that 
layer-by-layer the CNN is able to learn a representation that is invariant to real world
 factors such as light variation, view point distortion, as described in \cite{goodfellow2009measuring}, so that the representation obtained
  at the end of the network is invariant to all factors appearing in real images, and thus 
  comparable to the prototype.  

We adjust the traditional CNN pipeline to map both the input and prototypes to a
common feature space with the end goal of minimizing the final recognition error. The idea of
a common space for recognizing the instances by matching them to their
correct prototype is shown in Figure~\ref{fig:scheme}. For current
experiments, this common feature space is defined preemptively by the
prototypes of classes in context. Arguably, the prototypical templates, unaffected by noise
and distortion, are qualified to define an optimal embedding for maximum discrimination of classes.

The use of a joint embedding space lends the proposed model an interesting possibility 
of applying it to recognize new classes not present at the training stage. This aligns 
the approach within the areas of zero and one-shot learning. These areas pursue to 
emulate the ability of human beings to extrapolate and draw inference on test samples 
only from a description, or a single instance per class. Indeed, this is a faculty humans 
own, for example when assimilating and recognizing a new character such as \euro, after 
being presented with one instance. 



 

This paper makes the following contributions : (a) development of a CNN that is
able to use prototypical information to guide its learning process, (b) its application to classification
tasks presenting a boost in overall performance, (c) establishment of a new
benchmark in logo recognition (on Belga logo dataset), and (d) the seamless
application of the proposed model in zero-shot learning scenarios, given the
prototypical information of new classes at run time.


The paper is organized as follows. In Section~\ref{sec:Related-Works} we review related work. 
Section \ref{sec:Proposed-Approach} discusses the proposed approach. Sections
\ref{sec:Implementation-Details} and \ref{sec:Experiments} successively present the implementation details and
our experimental findings. Finally, Section~\ref{sec:Conclusion} concludes the
paper with a discussion about the presented work and a description of future
directions.

\section{\label{sec:Related-Works}Related Works}
Traditional computer vision approaches for classification do not take into
account the relationships there may be between the different output classes.
Arguably, if these relationships were available as side information, they could
be exploited to improve recognition performance.

Recent work focuses on taking advantage of this side information.
A considerable effort has been advocated to attribute learning. In this case,
side information takes the form of a high level description of each class as a list of attributes.
These attributes are often available in real datasets as tags, and have been
popularized within the research community thanks to datasets such as
\cite{Farhadi2009,lampert2009learning,patterson2012sun}.
Another side information that has recently been exploited by several works
\cite{socher2013zero_csm,frome2013devise,norouzi2013zero} is the semantic vector
representation of the name of each class. A semantic space of words can be
learned from a large corpus of text in an unsupervised way, so that
words are mapped to an Euclidean space in which the distance between vectors
depends on the semantic closeness of the words they represent. The vectors corresponding to the 
names of the classes can then be utilized as side information.

The availability of this side information about the relationship between classes
has led to the development of zero-shot learning, that is, the challenge of
identifying a class at test time without ever having seen samples of that during
training. Over the past few years, this idea has spurred much success,
using both attributes
\cite{pala_geoff2009zsl,akata2013label,romera2015embarrassingly,lampert2009learning},
and word embeddings \cite{norouzi2013zero,socher2013zero_csm}.

The developed approaches vary in the way knowledge is transferred from the training classes 
to the new classes. In \cite{Lampert2014,Suzuki2014} this transfer is done by
means of a cascaded probabilistic framework which determines the most likely class. One drawback of
probabilistic methods is that they make independence assumptions that do not
usually hold in practice. An alternative strategy which bypasses this drawback
has been recently exploited in
\cite{akata2013label,romera2015embarrassingly,weston2011wsabie}, where the
proposed model learns a linear embedding from both instances and attributes to a common space.
This can be seen as a two-layer model that connects the input images to class
labels through a layer containing attribute information. The weights connecting
the input space to the embedding space are learned to minimise the final
classification loss. Our proposed approach builds on this idea, although it presents two
significant differences. Firstly, the side information used consists of a
visual prototype for each class. Secondly, the mapping function from input to
embedding space is not linear, but modeled using a deepnet pipeline.

Another related area is that of one-shot learning
\cite{bart2005cross,lake2011one,fei2006one}. Similar to zero-shot learning, the objective here is
to transfer the knowledge learned at training stage to distinguish new classes.
The difference is that the information given to the model about the new classes
consists in one, or very few, instances.
One-shot learning is useful in image retrieval, where given an image as a query,
the model returns items that are similar \cite{seanBell2015}. Our work can be considered within this area, with the peculiarity that in
our framework the instance provided to the model is a very special one: it is a prototype. 
In fact, in our model the representation of the prototypes and input images could
be completely different (e.g. having different image size).

\section{\label{sec:Proposed-Approach}Proposed Approach}

In the usual image classification setup, given training samples of form
$(x, y)$, where $x \in \mathbb{R}^d$ is an image and $y \in
\left\{1,\ldots,C\right\}$ is the class label of the image, a classifier $h : \mathbb{R}^d \to \{1,\ldots,C\}$ 
is learned to predict the label of an unseen image $x$ as $\hat{y}$.

If we apply a regular $L$-layer
CNN to this problem, the function that is learned takes the following form:

\begin{equation}
\hat{y}=\operatornamewithlimits{argmax}_{c\in\{1,\ldots,C\}}\enspace
s\left(f_{L}\left(f_{L-1}\left(\ldots f_{2}\left(f_{1}\left(x;\theta_{1}\right);
\theta_{2}\right)\ldots;\theta_{L-1}\right);\theta_{L}\right)\right)_c.\label{eq:CNN}
\end{equation}

Here, $f_{l}$, for $l\in\left\{ 1,\ldots ,L\right\} $ represents the function
(e.g. convolution, pooling) applied at layer $l$, and $\theta_{l}$ denotes its
learnable parameters, if any. The last function $f_L$ maps its inputs to
$\mathbb{R}^C$. Finally, $s(.) : \mathbb{R}^C \to [0, 1]^C$ represents the
softmax activation function operating on a vector $z$, as follows: 

\begin{center}
$s(z)_{c}=\frac{\exp\left(z_{c}\right)}{\overset{C}{\underset{j=1}{\sum}}\exp\left(z_{j}\right)}$,
for $c\in\left\{ 1,\ldots ,C\right\} $,
\par\end{center}
where subscripts denote the elements of a vector.

During training, learnable weights $\theta_1, \theta_2, \ldots, \theta_L$ of the model are adjusted 
by backpropagating the negative log-likelihood loss over the ground truth label
${y}$ of a sample $x$, defined as follows:
\begin{center}
$\operatorname{loss}(x; \theta_1, \theta_2, \ldots, \theta_L) =
-\log(s(z)_{{y}}).$
\end{center}

The CNN represented by Equation~\eqref{eq:CNN} does not account for prior
information regarding prototypes of the classes. In order to introduce our
approach, let us assume that a prototype template image $p_{c}$ for each class
$c \in \{1,\ldots,C\}$ is provided. The proposed approach is based upon fixing the parameters of
the last layer of the CNN as a function of the prototype templates $p_c$, given by 
$\phi(p_{c})\in\mathbb{R}^{k}$, for some integer $k$, with $\|\phi(p_c)\|_2$ being constant for all $c
\in \{1,\ldots,C\}$. In practice, $\phi$ can be a feature extractor for the
template $p_c$; for instance, $\phi(p_c)$ can be a $k$-dimensional normalized HOG feature extracted from the
prototypical image $p_c$.

More specifically, we set $f_L : \R^k \to \R^C : f_L(v)_c = \langle\phi(p_c), v\rangle$, 
where $v$ denotes the activations fed into layer $L$ for a certain input
image, the subscript denotes vector elements and $\langle .,.\rangle$ denotes the usual dot product in $\R^k$. 
Since $\|\phi(p_c)\|_2$ is constant, when $c$ is varied for a fixed $v$,
$f_L(v)_c = \langle\phi(p_c), v\rangle$ attains the highest value for the
$\phi(p_c)$ closest to $v$ in the $k$-dimensional feature space.

The modified network can now be described using the following formula:
\begin{equation}
\hat{y}=\operatornamewithlimits{argmax}_{c \in \{1,\ldots,C\}}\enspace
s\left(f_{L}\left(f_{L-1}\left(\ldots
f_{2}\left(f_{1}\left(x\right)\right)\ldots\right)\right)\right)_c =
\operatornamewithlimits{argmax}_{c\in\{1,\ldots,C\}} \langle\phi(p_c),
\psi(x)\rangle,
\label{eq:CNN2}
\end{equation}
where $\psi(.)$ and $\phi(.)$ represent the projections of input images and
output labels into the joint feature space, respectively. An interpretation of this approach is that 
the learnable part of the network,
$\psi : \R^{d} \to \R^{k}: \psi=f_{L-1}\circ\ldots\circ f_{1}$,
learns a non-linear mapping from the original images to a $k$-dimensional
latent space, which in this case is defined by the prototypes. 
\begin{figure}
  \centering
  \includegraphics[width=0.95\columnwidth]{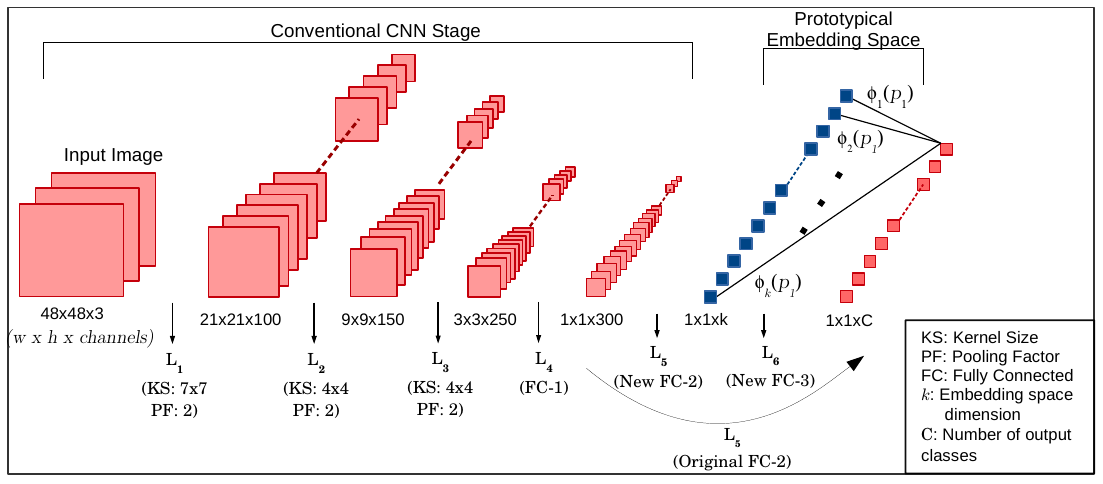}
\caption{\label{fig:network_arch}Network architecture with the introduction
  of prototypical priors. In the current experiments, $k$-dimensional HoG
  features extracted over the prototypical templates are used to define the
  common embedding space.}
\end{figure}
This space contains both the projections of an input image, $\psi(x)$, and the
prototype templates, $\phi(p_{c})$ for all $c\in\left\{ 1,\ldots ,C\right\} $,
such that the similarity between each instance-prototype pair can be computed by means of an inner product.
The use of softmax-loss function leads to a discriminative way to encourage the inner product 
between $\psi(x)$ and $\phi(p_{c})$ to be high if instance $x$ belongs to class
$c$, and to be low otherwise. Thus, we introduce prior
information about the classes directly into the network with the aspiration that the remaining parameters
will adapt themselves to accommodate the fixed last layer in the learning process.



Note that, unlike in other works such as \cite{pala_geoff2009zsl,norouzi2013zero}, at test
time the inference process is exactly the same as in any other CNN. There is no need to 
perform explicit calculations about distances in the embedding space.

This framework easily allows for using new prototypes after
the training stage is finished. This is done by replacing, or adding to the last
layer new weights according to the new prototypes. The resultant network is
potentially capable of distinguishing the new classes because the invariances
learned in $\psi$ are conceivably common to all classes.

In the given framework, both functions $\psi$ and $\phi$ can be learned.
However, for the purpose of current research, we focus on the
case where $\psi$ is learned as part of the traditional CNN pipeline, while $\phi$ is fixed by 
a prescribed function, such as HoG transform.


%
%

\section{\label{sec:Implementation-Details}Implementation Details}
We now detail the architecture of our deep network used to implement the ideas described above. 
The first stage of our network consists of a CNN to enable learning of image features starting from 
original RGB patches of $48\times48$ (size suitable for both traffic-sign and
logo samples in experimental datasets). 

The configuration, as presented in \textit{red} (\textit{light} for grayscale)
in Figure ~\ref{fig:network_arch}, is the same from {\cite{multicolumnarDNN}}
with the exception of a dropout layer after $L_{5}$.
As in a traditional CNN designed for classification, the last few layers are fully-connected, and the 
network is terminated with a layer having the same number of activations as the number of classes $C$. 
A softmax function is applied to the last layer to obtain a probability distribution over the output class 
labels. 

In the proposed approach, prototypical information is introduced by wedging a layer before the output 
layer, fully connected to the $C$ output neurons using the fixed weights $\phi(p_c)\in \mathbb{R}^k$ for 
all $c\in\left\{ 1,\ldots,C\right\}$. The new layer and its connections are
shown in \textit{blue} (\textit{dark} for grayscale) in
Figure ~\ref{fig:network_arch}.
Thus, the $k\times C$ weight matrix for the last fully connected layer $f_{L}$
is defined as a set of $k\times1$ vectors $\phi(p_c)$ one for each
$c\in\{{1,\ldots, C}\}$. In Figure ~\ref{fig:network_arch}, we use $\phi_1(p_c),
\phi_2(p_c), \ldots,  \phi_k(p_c)$ to represent the elements of the $k$-dimensional vector $\phi(p_c)$.

In the current work, we fix the embedding space using $k$-dimensional
normalized histograms of oriented gradients (HoG) {\cite{dalal2005histograms}} features
extracted from the prototypical templates. The prototypical images are all resized to a fixed size $s\times s$.
The HOG features are extracted using the standard \textit{extractHoGFeatures} function provided with Matlab. 
In our experiments, we make use of an empirically selected cell size $c=10$, block size $b=2$,  
overlap factor $o=1$ and a bin count $n=12$, for $s=100$, which yields $3888$-dimensional HoG features.
%


\section{\label{sec:Experiments}Experiments}
We explore the above idea of introducing prototypical information during deep learning phase for 
two end goals: (a) Improvement in overall classification performance when all classes are seen during 
training, (b) Improvement in classification performance over unseen classes,
i.e., in a zero-shot learning scenario.

\begin{figure}
  \centering
  \includegraphics[width=0.8\columnwidth]{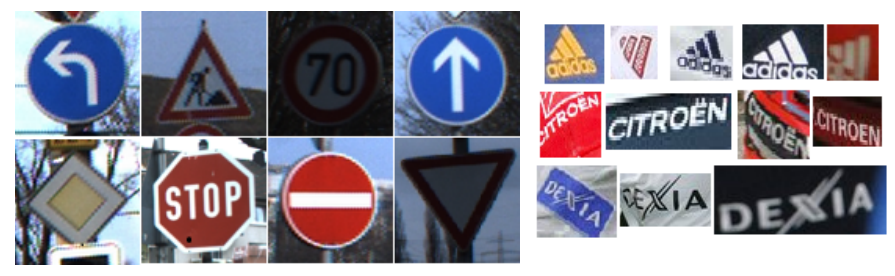}
\caption{\label{fig:sampleimages}Sample images of traffic signs (left) show view
point distortion, illumination variation and background clutter while logo
images (right) additionally contain non-planar distortions and high
self-occlusion. }
\end{figure}

\subsection{\label{}Datasets}
To analyse the generalisability of the proposed approach we evaluate it on two
separate datasets as described below.

\textit{Traffic Sign Dataset:} We use the German Traffic Sign Recognition
benchmark \cite{stallkamp2012manvscomp}, hereafter referred to as D1. This dataset has a substantial 
sample base of more than 50,000 images spread over 43 traffic sign classes. 
The dataset is divided into 39,209 training samples and 12,630 test samples. 
For experimental purpose, we randomly split the test data into validation and 
test sets of 6,315 samples each. We fine crop the samples using the information provided with the 
dataset. No additional distortion (such as scaling, rotation) is applied at training or testing time. 

\textit{Brand Logo Dataset:} We use the Belga Logos dataset
{\cite{belgajoly2009logo}}, hereafter referred to as D2.
The dataset contains bounding box annotations for 37 logo categories collected from across 10,000 
real images. Out of a total of 9,841 logo samples, 2,697 are marked as `OK' for their ability to be 
recognizable without the image context. We use a subset of 10 logo classes (out of the 37), for 
which the total number of samples per class is at least 100. We set aside 20\% of the samples from each class for validation, and 20\% for testing.
 
Sample images from both the datasets are as shown in Figure
~\ref{fig:sampleimages}.
\begin{table}
\begin{centering}
\begin{tabular}{|c|c|c|}
\hline 
 \shortstack{Dropout \\ Factor}& \shortstack{Case 1 \\ Test accuracy
 (\% )} & \shortstack{Case 2 \\ Test accuracy in (\%
)}\tabularnewline
\hline 
\hline 
0.5 & 96.60 & 97.98\tabularnewline
\hline 
0.6 & 97.18 & 97.53\tabularnewline
\hline 
0.65 & 97.48 & 97.74\tabularnewline
\hline 
\end{tabular}
\vspace{6pt}
\protect\caption{\label{zslconfigs} Consistent boost in classification
performance across different configurations experimented for dataset D1. The
test accuracy compares to \cite{multicolumnarDNN} in the case of no use of data
augmentation during training.}
\end{centering}
\end{table}

\begin{table}
\begin{centering}
\begin{tabular}{|c|c|c|}
\hline 
Dataset & \shortstack{Case 1 \\ Test accuracy (\% )}  
& \shortstack{Case 2 \\ Test accuracy in (\%
)}\tabularnewline
\hline 
\hline 
D1 & 97.48 & 97.98\tabularnewline
\hline 
D2 & 93.48 & 93.57\tabularnewline
\hline 
\end{tabular}
\vspace{6pt}
\protect\caption{\label{classperformance} Overall improvement in classification
performance with the use of prototypical information}
\end{centering}
\end{table}
\subsection{\label{}Results}
\subsubsection{\label{overall}Overall Recognition Performance}
In this setup, all the classes are treated as seen. Classification
results on dataset D1 with comparable configurations of conventional (Case 1)
and proposed (Case 2) deepnet pipeline are shown in Table \ref{zslconfigs}. For
the 3 different configurations, dropout-factor of layer after $L_5$ is varied to be 0.5, 0.6 and 0.65
respectively. 

Top results on D1 and D2, without (Case 1) and with (Case 2) the use of
prototypical information, are shown in Table \ref{classperformance}. For dataset D1, test performance 
without prototypical information is comparable to that presented in
\cite{multicolumnarDNN} for the case when no additional data augmentation technique is employed. Inclusion
of prototypical embedding boosts the performance by 0.5\% leading
to an almost 20\% reduction in the error rate. On dataset D2, the proposed
approach gives a comparable performance to, if not better than, the
baseline.
A possible explanation could be that logo samples display heavy self
occlusion, perspective distortion and general lack of visual quality.

\textit{Additional findings:} For both the datasets, we experimented with
grayscale as well as colored (RGB) prototypes. Models using prototypical features extracted from 
colored templates consistently performed lower (by an average margin of 0.1\%)
compared to those using the same features obtained from grayscale templates. 

This suggests that while color coding may be useful in
garnering visual attention, it may not be quintessential for distinguishing the
classes. For traffic sign dataset D1, 12 out of 43 classes
are \textit{Prohibitory} traffic signs with a consistent circular red and white
color coding, while 8 are \textit{Mandatory} signs with a uniform circular blue
and white color coding. Evidently, the main discrimination quotient in traffic signs is added by
the inset depiction. 
On the other hand, for logo dataset D2, samples show significant color
variation within a single class,  as shown in ~\ref{fig:sampleimages}, which
renders the color information quite irrelevant.
\begin{figure}
  \centering
  \includegraphics[width=0.7\columnwidth]{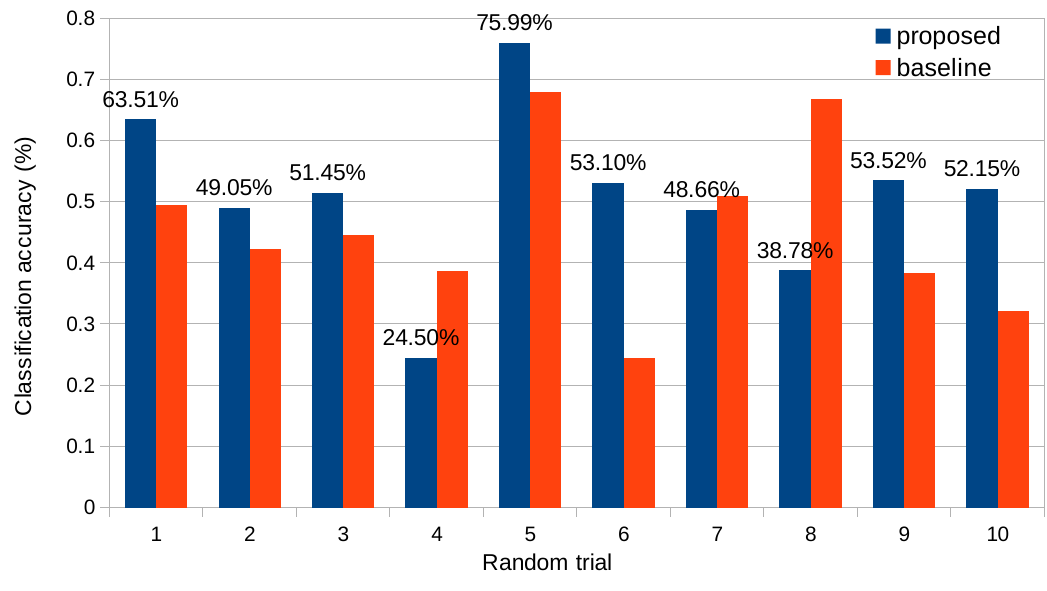}
\caption{\label{fig:comparisonD1}Recognition performance on unseen classes of
dataset D1 compared across proposed and baseline \cite{norouzi2013zero} approach}
\end{figure}

\begin{figure}
  \centering
  \includegraphics[width=0.7\columnwidth]{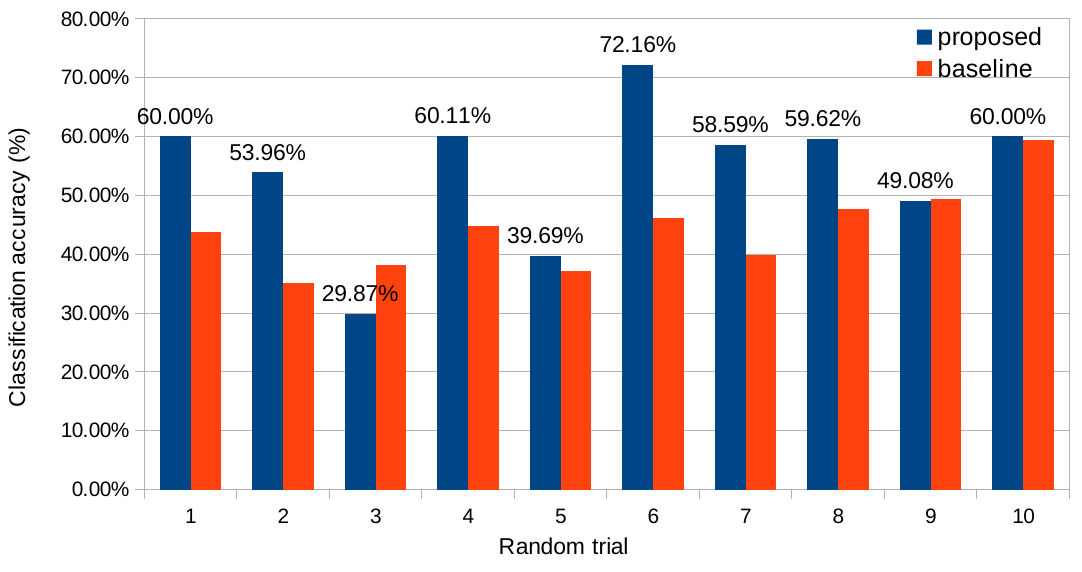}
\caption{\label{fig:comparisonD2}Recognition performance on unseen classes of
dataset D2 compared across proposed and baseline \cite{norouzi2013zero}
approach}
\end{figure}
\subsubsection{\label{}Zero-Shot Learning}
\textit{Data setup:}  The 43 classes of dataset D1 are
divided into 33 seen classes (denoted by the set of classes $C_s$), and 10 unseen classes (denoted by the 
set $C_u$). Samples from classes in $C_s$ are used for 
training the model while the remaining 10 classes in $C_u$ are used for testing the model. During test time, all $c \in C_u$ form the output 
label set, that is, the network could predict any label from $C_u$.

Similarly for D2, 10 classes are divided into 7 seen classes (set $C_s$) and 3
unseen classes (set $C_u$). Samples with class labels in $C_s$ are used for
training the model and the 3 classes in $C_u$ are used for testing.

\textit{Comparison:} We compare our approach with the method
of convex combination of embedding vectors, as
discussed in \cite{norouzi2013zero}. In this, new unseen class samples are represented as
weighted combinations of vector embeddings $\phi(p_c)$ of seen classes $c \in
C_s$, where the weights are the probabilistic output of the softmax layer. Top
$T$-predictions are combined to yield the feature representation, where $T$ is a
hyperparameter that can be tuned by means of a validation process. These
representations are compared in the vector space defined by
$\phi(p_c)$, where $c \in C_u$.
The class of the input sample is inferred to be the class of the nearest
prototype in this space.

\textit{Findings and discussion:} We make 10 random selections of $C_s$ and
$C_u$. For the proposed approach, the prototypical
representations of $\phi(p_c),\,c\in C_s$ are used during training, while these are replaced 
by $\phi(p_c),\,c\in C_u$ during testing. The validation
hyperparameter $T$ of \cite{norouzi2013zero} is set to the total
number of seen classes $C_s$ while the proposed approach simply
validates against a set-aside sample set over all the seen classes.
The classification results for unseen classes on datasets D1 and D2 
are compared in Figures ~\ref{fig:comparisonD1} and ~\ref{fig:comparisonD2}
respectively.
The proposed approach outperforms \cite{norouzi2013zero} with an average
accuracy gain of 5.48\% and 10.15\% on datasets D1 and D2 respectively. The
performance gain is statistically significant for D1 at a p-value of 5\% as well
as for D2 although with a p-value of 33\%.
Due to visual similarity, an unseen traffic sign can still be fairly well reproduced by the combination of
related prototypical templates as done in \cite{norouzi2013zero}. The major
benefit of proposed approach is evident in visually dissimilar logo categories
where the zero shot performance is considerably improved.

\begin{figure}[t]
  	\begin{center}
  	\includegraphics[trim = 0mm 2mm 0mm 0.1mm, clip,scale=0.55]{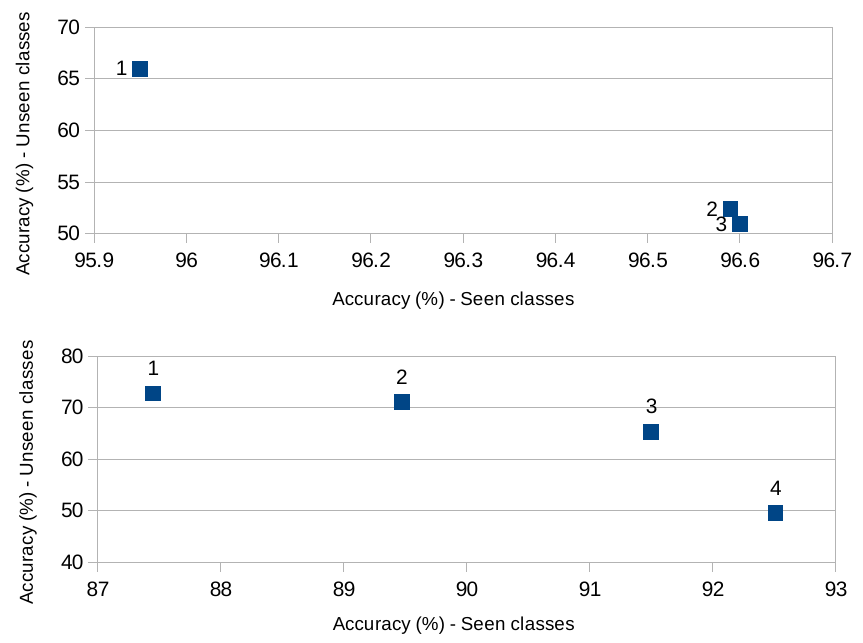}
  	\end{center}
  \caption{\label{tradeoff}Performance trade-off curve 
for seen and unseen classes over a certain trial of D1 and D2 respectively}
\end{figure}
In the approach of \cite{norouzi2013zero}, training and
validation are disconnected steps. The CNN can be trained for maximum
performance only on $C_s$. At validation time parameter $T$, that defines the
number of seen classes used for drawing inference, provides little flexibility for tuning the performance 
on $C_u$. On the contrary, our model can be fine-tuned either for unseen
or seen class performance by validating against the appropriate set. In
the current experiments, we validate against a sample set collected over the
seen classes $C_s$, however it can also contain samples from a few unseen
classes $c\in C_u$ marked as validation classes. CNN training is carried out as
before to get a joint optimization for both seen and unseen class performance.

In our experimental experience we found that the performances of seen and unseen
classes are positively correlated in the initial stages of the training
procedure. However, this happens up to a point, beyond which both performances
appear to be negatively correlated (see Figure \ref{tradeoff} showing the performance
trade-off curve for seen and unseen classes over a certain trial of D1 and D2 respectively using
our approach). The above tests are carried out using a certain random
selection of 5 unseen classes for D1 and 2 unseen classes for D2.

\section{\label{sec:Conclusion}Conclusion}

In this paper we showed that visual prototypes can be successfully used as side information 
to aid the learning process in traditional classification setup as well as for
zero-shot learning.

We proposed a method for integrating prototypical information in the
successful deep learning framework. Using a conventional CNN stage, the input projection
function that maps input images to a joint prototypical space can be learned for 
maximum similarity between a real-world instance and its prototype, while minimising the
end recognition loss. In the current research, this embedding space is
preemptively fixed by the choice of prototypical representation while the input mapping is
learnable as a complex non-linear function. More generally, however, both the
input and output embeddings can be learned as an end-to-end deepnet pipeline. We
plan to explore this as part of our future work.

As observed on two different datasets of traffic signs and brand logos, results
of the proposed approach are highly promising.
Regarding its application to regular object recognition, we can conclude that constraining the
network to incorporate the given prototypes does not hamper, but on the contrary
 improves the classification performance. With regard to zero-shot learning, our model shows better results than a state-of-the-art competitor~\cite{norouzi2013zero}. Furthermore,  
our model can be flexibly trained for the required trade-off between seen and unseen class performance, 
and inference on new unseen classes simply involves adding their prototypical
information at test time.


\begin{thebibliography}{23}
\providecommand{\natexlab}[1]{#1}
\providecommand{\url}[1]{\texttt{#1}}
\expandafter\ifx\csname urlstyle\endcsname\relax
  \providecommand{\doi}[1]{doi: #1}\else
  \providecommand{\doi}{doi: \begingroup \urlstyle{rm}\Url}\fi

\bibitem[Akata et~al.(2013)Akata, Perronnin, Harchaoui, and
  Schmid]{akata2013label}
Zeynep Akata, Florent Perronnin, Zaid Harchaoui, and Cordelia Schmid.
\newblock Label-embedding for attribute-based classification.
\newblock In \emph{Computer Vision and Pattern Recognition (CVPR), 2013 IEEE
  Conference on}, pages 819--826. IEEE, 2013.

\bibitem[Bart and Ullman(2005)]{bart2005cross}
Evgeniy Bart and Shimon Ullman.
\newblock Cross-generalization: Learning novel classes from a single example by
  feature replacement.
\newblock In \emph{Computer Vision and Pattern Recognition, 2005. CVPR 2005.
  IEEE Computer Society Conference on}, volume~1, pages 672--679. IEEE, 2005.

\bibitem[Cire{\c{s}}an et~al.(2012)Cire{\c{s}}an, Meier, Masci, and
  Schmidhuber]{multicolumnarDNN}
Dan Cire{\c{s}}an, Ueli Meier, Jonathan Masci, and J{\"u}rgen Schmidhuber.
\newblock Multi-column deep neural network for traffic sign classification.
\newblock \emph{Neural Networks}, 32:\penalty0 333--338, 2012.

\bibitem[Dalal and Triggs(2005)]{dalal2005histograms}
Navneet Dalal and Bill Triggs.
\newblock Histograms of oriented gradients for human detection.
\newblock In \emph{Computer Vision and Pattern Recognition, 2005. CVPR 2005.
  IEEE Computer Society Conference on}, volume~1, pages 886--893. IEEE, 2005.

\bibitem[Farhadi et~al.(2009)Farhadi, Endres, Hoiem, and Forsyth]{Farhadi2009}
a.~Farhadi, I.~Endres, D.~Hoiem, and D.~Forsyth.
\newblock {Describing objects by their attributes}.
\newblock \emph{2009 IEEE Conference on Computer Vision and Pattern
  Recognition}, pages 1778--1785, June 2009.
\newblock \doi{10.1109/CVPR.2009.5206772}.

\bibitem[Fei-Fei et~al.(2006)Fei-Fei, Fergus, and Perona]{fei2006one}
Li~Fei-Fei, Robert Fergus, and Pietro Perona.
\newblock One-shot learning of object categories.
\newblock \emph{Pattern Analysis and Machine Intelligence, IEEE Transactions
  on}, 28\penalty0 (4):\penalty0 594--611, 2006.

\bibitem[Frome et~al.(2013)Frome, Corrado, Shlens, Bengio, Dean, Mikolov,
  et~al.]{frome2013devise}
Andrea Frome, Greg~S Corrado, Jon Shlens, Samy Bengio, Jeff Dean, Tomas
  Mikolov, et~al.
\newblock Devise: A deep visual-semantic embedding model.
\newblock In \emph{Advances in Neural Information Processing Systems}, pages
  2121--2129, 2013.

\bibitem[Goodfellow et~al.(2009)Goodfellow, Lee, Le, Saxe, and
  Ng]{goodfellow2009measuring}
Ian Goodfellow, Honglak Lee, Quoc~V Le, Andrew Saxe, and Andrew~Y Ng.
\newblock Measuring invariances in deep networks.
\newblock In \emph{Advances in neural information processing systems}, pages
  646--654, 2009.

\bibitem[Joly and Buisson(2009)]{belgajoly2009logo}
Alexis Joly and Olivier Buisson.
\newblock Logo retrieval with a contrario visual query expansion.
\newblock In \emph{Proceedings of the 17th ACM international conference on
  Multimedia}, pages 581--584. ACM, 2009.

\bibitem[Krizhevsky et~al.(2012)Krizhevsky, Sutskever, and
  Hinton]{Krizhevsky/nips2012}
Alex Krizhevsky, Ilya Sutskever, and Geoffrey~E. Hinton.
\newblock Imagenet classification with deep convolutional neural networks.
\newblock In F.~Pereira, C.J.C. Burges, L.~Bottou, and K.Q. Weinberger,
  editors, \emph{Advances in Neural Information Processing Systems 25}, pages
  1097--1105. Curran Associates, Inc., 2012.

\bibitem[Lake et~al.(2011)Lake, Salakhutdinov, Gross, and
  Tenenbaum]{lake2011one}
Brenden~M Lake, Ruslan Salakhutdinov, Jason Gross, and Joshua~B Tenenbaum.
\newblock One shot learning of simple visual concepts.
\newblock In \emph{Proceedings of the 33rd Annual Conference of the Cognitive
  Science Society}, volume 172, 2011.

\bibitem[Lampert et~al.(2009)Lampert, Nickisch, and
  Harmeling]{lampert2009learning}
Christoph~H Lampert, Hannes Nickisch, and Stefan Harmeling.
\newblock Learning to detect unseen object classes by between-class attribute
  transfer.
\newblock In \emph{Computer Vision and Pattern Recognition, 2009. CVPR 2009.
  IEEE Conference on}, pages 951--958. IEEE, 2009.

\bibitem[Lampert et~al.(2014)Lampert, Nickisch, and Harmeling]{Lampert2014}
Christoph~H Lampert, Hannes Nickisch, and Stefan Harmeling.
\newblock Attribute-based classification for zero-shot visual object
  categorization.
\newblock \emph{IEEE transactions on pattern analysis and machine
  intelligence}, 36\penalty0 (3):\penalty0 453--65, March 2014.
\newblock ISSN 1939-3539.
\newblock \doi{10.1109/TPAMI.2013.140}.

\bibitem[Lee et~al.(2009)Lee, Grosse, Ranganath, and Ng]{Lee/icml2009}
Honglak Lee, Roger Grosse, Rajesh Ranganath, and Andrew~Y Ng.
\newblock Convolutional deep belief networks for scalable unsupervised learning
  of hierarchical representations.
\newblock In \emph{Proceedings of the 26th Annual International Conference on
  Machine Learning}, pages 609--616. ACM, 2009.

\bibitem[Norouzi et~al.(2013)Norouzi, Mikolov, Bengio, Singer, Shlens, Frome,
  Corrado, and Dean]{norouzi2013zero}
Mohammad Norouzi, Tomas Mikolov, Samy Bengio, Yoram Singer, Jonathon Shlens,
  Andrea Frome, Greg~S Corrado, and Jeffrey Dean.
\newblock Zero-shot learning by convex combination of semantic embeddings.
\newblock \emph{arXiv preprint arXiv:1312.5650}, 2013.

\bibitem[Palatucci et~al.(2009)Palatucci, Pomerleau, Hinton, and
  Mitchell]{pala_geoff2009zsl}
Mark Palatucci, Dean Pomerleau, Geoffrey~E Hinton, and Tom~M Mitchell.
\newblock Zero-shot learning with semantic output codes.
\newblock In \emph{Advances in neural information processing systems}, pages
  1410--1418, 2009.

\bibitem[Patterson and Hays(2012)]{patterson2012sun}
Genevieve Patterson and James Hays.
\newblock Sun attribute database: Discovering, annotating, and recognizing
  scene attributes.
\newblock In \emph{Computer Vision and Pattern Recognition (CVPR), 2012 IEEE
  Conference on}, pages 2751--2758. IEEE, 2012.

\bibitem[Romera-Paredes et~al.(2015)Romera-Paredes, OX, and
  Torr]{romera2015embarrassingly}
Bernardino Romera-Paredes, ENG OX, and Philip~HS Torr.
\newblock An embarrassingly simple approach to zero-shot learning.
\newblock In \emph{Proceedings of The 32nd International Conference on Machine
  Learning}, pages 2152--2161, 2015.

\bibitem[Sean~Bell(2015)]{seanBell2015}
Kavita~Bala Sean~Bell.
\newblock Learning visual similarity for product design with convolutional
  neural networks.
\newblock \emph{ACM Transactions on Graphics (SIGGRAPH 2015)}, 2015.

\bibitem[Socher et~al.(2013)Socher, Ganjoo, Manning, and
  Ng]{socher2013zero_csm}
Richard Socher, Milind Ganjoo, Christopher~D Manning, and Andrew Ng.
\newblock Zero-shot learning through cross-modal transfer.
\newblock In \emph{Advances in Neural Information Processing Systems}, pages
  935--943, 2013.

\bibitem[Stallkamp et~al.(2012)Stallkamp, Schlipsing, Salmen, and
  Igel]{stallkamp2012manvscomp}
Johannes Stallkamp, Marc Schlipsing, Jan Salmen, and Christian Igel.
\newblock Man vs. computer: Benchmarking machine learning algorithms for
  traffic sign recognition.
\newblock \emph{Neural networks}, 32:\penalty0 323--332, 2012.

\bibitem[Suzuki et~al.(2014)Suzuki, Sato, Oyama, and Kurihara]{Suzuki2014}
Masahiro Suzuki, Haruhiko Sato, Satoshi Oyama, and Masahito Kurihara.
\newblock {Transfer learning based on the observation probability of each
  attribute}.
\newblock \emph{2014 IEEE International Conference on Systems, Man, and
  Cybernetics (SMC)}, pages 3627--3631, October 2014.
\newblock \doi{10.1109/SMC.2014.6974493}.

\bibitem[Weston et~al.(2011)Weston, Bengio, and Usunier]{weston2011wsabie}
Jason Weston, Samy Bengio, and Nicolas Usunier.
\newblock Wsabie: Scaling up to large vocabulary image annotation.
\newblock In \emph{IJCAI}, volume~11, pages 2764--2770, 2011.

\end{thebibliography}

\end{document}